\documentclass[10pt,a4paper,logo]{paper}

\usepackage[all]{hypcap}

\usepackage[authoryear, round]{natbib}

\usepackage{nicefrac}       
\usepackage{multirow}
\usepackage{cleveref}
\usepackage{dsfont}
\usepackage{makecell}
\usepackage{subcaption}
\usepackage{bbm}
\usepackage{wrapfig}
\usepackage{mathtools,xparse}
\usepackage{array}
\usepackage{arydshln}
\usepackage{scalerel}
\usepackage{tablefootnote}
\usepackage{tocloft}
\usepackage{adjustbox}
\usepackage{setspace}
\usepackage[most,skins,theorems]{tcolorbox}

\newtcolorbox{promptbox}{
  colback=gray!5,
  colframe=gray!70,
  boxrule=0.8pt,
  arc=2mm,
  left=10pt,
  right=10pt,
  top=10pt,
  bottom=10pt,
  width=\linewidth,
  fontupper=\ttfamily\footnotesize,
  enhanced,
}

\newboolean{showsection}
\setboolean{showsection}{true}
\makeatletter
\@namedef{ver@everyshi.sty}{}
\makeatother

\definecolor{custom_green}{rgb}{0.0, 0.5, 0.0}
\definecolor{custom_red}{rgb}{1.0, 0.01, 0.24}
\definecolor{custom_blue}{HTML}{C9DAF7}
\definecolor{custom_purple}{HTML}{D9D1E9}
\definecolor{title_blue}{HTML}{204899}
\definecolor{cite_blue}{HTML}{044dc1}
\definecolor{cite_purple}{HTML}{7406a7}

\hypersetup{
    colorlinks = true,
    citecolor = {cite_blue},
    linkcolor = {cite_purple},
    urlcolor = {cite_blue},
}

\definecolor{blanchedalmond}{rgb}{1.0, 0.92, 0.8}
\definecolor{carmine}{rgb}{0.59, 0.0, 0.09}
\definecolor{lightblue}{rgb}{0.22,0.45,0.70}%

\renewcommand{\mathbf}{\boldsymbol}

\makeatletter
\def\Ddots{\mathinner{\mkern1mu\raise\p@
\vbox{\kern7\p@\hbox{.}}\mkern2mu
\raise4\p@\hbox{.}\mkern2mu\raise7\p@\hbox{.}\mkern1mu}}
\makeatother

\numberwithin{equation}{section}

\definecolor{amaranth}{rgb}{0.9, 0.17, 0.31}
\definecolor{antiquebrass}{rgb}{0.8, 0.58, 0.46}
\definecolor{antiquefuchsia}{rgb}{0.57, 0.36, 0.51}
\definecolor{chromeyellow}{rgb}{0.31, 0.47, 0.26}

\newcommand{\1}{\mathds 1}

\usepackage{amsmath,amsfonts,bm}

\def\eqref#1{equation~\ref{#1}}

\def\1{\bm{1}}

\DeclareMathAlphabet{\mathsfit}{\encodingdefault}{\sfdefault}{m}{sl}
\SetMathAlphabet{\mathsfit}{bold}{\encodingdefault}{\sfdefault}{bx}{n}

\makeatletter
\def\mathcolor#1#{\@mathcolor{#1}}
\def\@mathcolor#1#2#3{%
  \protect\leavevmode
  \begingroup
    \color#1{#2}#3%
  \endgroup
}
\makeatother

\Crefformat{equation}{#2Eq.\;(#1)#3}
\Crefformat{figure}{#2Figure #1#3}
\Crefformat{assumption}{#2Assumption #1#3}
\Crefname{assumption}{Assumption}{Assumptions}

\usepackage{crossreftools}
\makeatletter
\renewcommand\footnoterule{%
  \kern 15\p@
  \hrule \@width 2in \kern 2.6\p@
  \vspace{4pt}
}
\makeatother

\title{OmniRetrieval: Unified Retrieval across~Heterogeneous~Knowledge~Sources}

\reportnumber{}

\author[1]{Jinheon Baek}
\author[1]{Soyeong Jeong}
\author[1]{Sangwoo Park}
\author[1]{Woongyeong Yeo}
\author[1]{Minki Kang}
\author[2]{\\Patara Trirat}
\author[2]{Heejun Lee}
\author[1,2]{Sung Ju Hwang}

\affil[1]{KAIST}
\affil[2]{DeepAuto.ai}

\correspondingauthor{Correspondence to: \email{\{jinheon.baek, sungju.hwang\}@kaist.ac.kr}.}

\begin{abstract}
Real-world information needs require access to structurally diverse knowledge sources, from unstructured text and relational tables to knowledge graphs and property graphs. Existing retrievers, however, operate over one source at a time under a fixed query language, leaving the broader landscape of available knowledge fragmented behind incompatible interfaces. A natural attempt at unification would collapse these sources into a shared space, but this erases the structural affordances (such as schemas, ontologies, compositional operators) that give each source its expressive power. Effective retrieval over diverse knowledge, therefore, requires not homogenization but an overarching layer that meets each source on its own terms. To achieve this, we present OmniRetrieval, a framework that takes any natural-language query, identifies appropriate knowledge sources, and dispatches source-native queries to their native execution engines. Across an extensive benchmark spanning 13 datasets and 309 distinct knowledge bases over text, relational, and graph-structured sources, OmniRetrieval exceeds single-source baselines, demonstrating that it can serve as a general-purpose interface to the heterogeneous sources while preserving the structural distinctions that make each source valuable. Our code is available at \url{https://github.com/JinheonBaek/OmniRetrieval}.
\end{abstract}

\begin{document}

\maketitle

\section{Introduction}

\begin{figure*}[h]
    \centering
    \vspace{-0.025in}
    \includegraphics[width=0.975\linewidth]{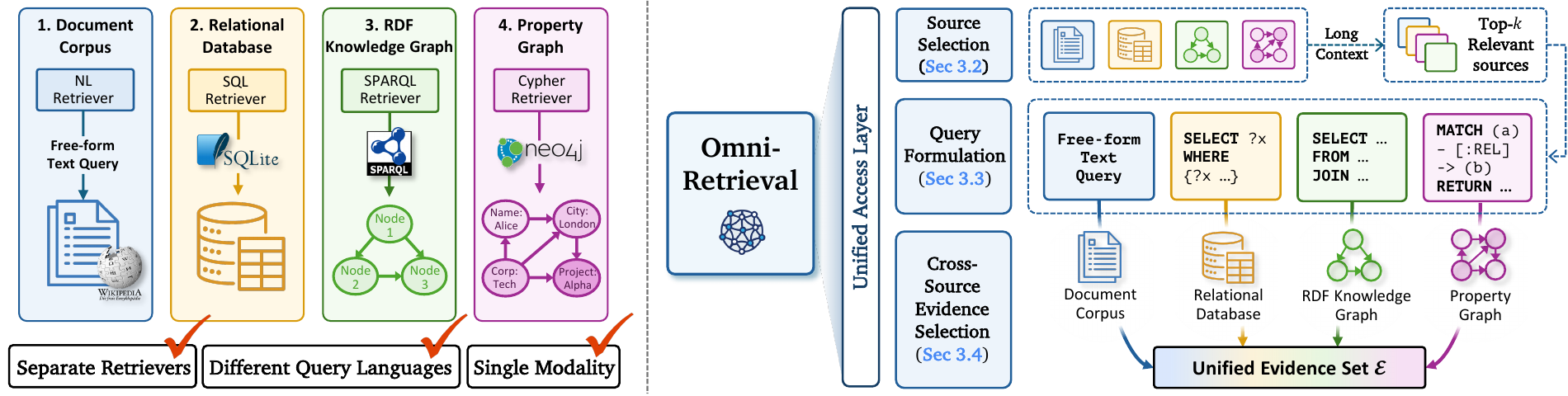}
    \vspace{-0.050in}
    \caption{Different knowledge sources offer distinct structural affordances and query languages (left); OmniRetrieval meets each on its own terms via source selection, native query formulation, and cross-source condensation (right).}
    \label{fig:concept}
\end{figure*}

The knowledge that answers a real-world question rarely lives in a single place, or in a single shape. A clinical question may be answered by a passage in a biomedical article~\citep{BEIR}; an enterprise question may require a join across normalized relational tables~\citep{Spider, Bird}; a factoid question about people, places, or events may resolve to a few triples in an encyclopedic knowledge graph~\citep{Freebase, Wikidata}; and a question about a supply chain or an academic collaboration network may turn on a multi-hop traversal of a labeled property graph~\citep{Text2Cypher}. In each case, the right answer is, in principle, retrievable, but only if one already knows which corpus to consult, which query language to write, and which execution engine to dispatch it to. The retrieval problem, then, is not merely to find relevant content within a source, but to navigate the structural heterogeneity that runs across sources.

Existing retrieval approaches, however, are typically designed for one source at a time. Specifically, document retrievers operate over an unstructured corpus and rank passages by similarity to a free-form query~\citep{BM25, DPR}; text-to-SQL systems target a single relational database and emit a single SQL dialect~\citep{Spider, Bird}; SPARQL or Cypher generators are likewise tied to a single graph backend and query language, with SPARQL for RDF stores and Cypher for labeled property graphs~\citep{Text2SPARQL, Text2Cypher}. As a consequence, even when a recent Large Language Model (LLM) can reason across evidence drawn from many kinds of sources~\citep{Claude3, Gemini-2.5, GPT-5}, the retrieval layer that feeds it cannot reach into them all, leaving the broader knowledge landscape out of reach.

A natural response is to collapse the silos themselves, by projecting every knowledge source into a shared representation, typically a single dense embedding space or a shared linearized text format~\citep{UniK, UDT, DiFaR}. However, this restores a uniform interface at a cost: the structural affordances that distinguish each source are flattened away, and what remains is a lossy projection with two consequences. First, the unified embeddings cluster by source type rather than by semantic content, a modality gap that biases retrieval toward sources resembling the query in form rather than those that actually answer it~\citep{UniversalRAG}. Second, it supports only the similarity matching, and the native query operations of each source are lost. Inspired by these, the right move, we argue, is the opposite of homogenization: keep each source on its own terms, and instead build a unifying \emph{access layer} above them.

In this work, we instantiate this view in \textbf{OmniRetrieval}, a framework that engages the knowledge sources relevant to a query, each through its own query language (Figure~\ref{fig:concept}). Specifically, given a query, OmniRetrieval identifies which of the available knowledge sources are relevant, and for each such source formulates an executable query in the corresponding native language, conditioning on whatever structural context the source exposes. These queries are then executed by the sources themselves, and when more than one source has been engaged, the resulting outputs are consolidated to keep those relevant to the question. Notably, as every source is reached on its own terms rather than through a shared representation, adding a new base is a matter of registration alone, with no shared encoder to retrain and no embedding space to redraw.

We then evaluate OmniRetrieval on an extensive benchmark that spans 309 distinct knowledge bases across four backend types (such as unstructured corpora, relational databases, RDF knowledge graphs, and labeled property graphs), drawn from 13 publicly available datasets~\citep{SimpleQuestions, Spider, LCQuAD2, BEIR, Bird, QALD10, Text2Cypher}. Across this suite, OmniRetrieval consistently exceeds single-source baselines constrained to operate within one modality, while also identifying the correct knowledge bases for each query with high accuracy and producing structurally valid queries in each native language. These results affirm that unified retrieval across distinct knowledge sources can be achieved by coordinating access through the native interfaces of each source.

\section{Method}

\subsection{Problem Formulation}

Let $q$ be a question from a user and $\mathcal{B} = \{b_1, \ldots, b_N\}$ be a pool of independently maintained knowledge sources. Each of these sources $b \in \mathcal{B}$ has its own native query language (such as SQL for a relational database, SPARQL for an RDF graph, Cypher for a labeled property graph, or free-form text for an unstructured corpus), its own execution engine $\texttt{Exec}(b, \hat{q})$ that accepts a native query $\hat{q}$ (written in that language) and returns a set of results, and an exposed structural context $c_b$ (such as a relational schema, an ontology, or a corpus descriptor) that any external caller can read in order to formulate an executable query against $b$. However, knowledge sources may differ arbitrarily in what they store and how they store it, where one may hold unstructured text, another normalized tables, and a third a labeled graph, and they return their results in correspondingly different forms.

The retrieval task is then to find and provide, for the question $q$, a set of evidence drawn from one or more sources in $\mathcal{B}$ that is relevant to $q$. Notably, a retrieval framework addressing this task should operationalize the selection of a subset $\mathcal{S} \subseteq \mathcal{B}$ of sources to engage, the formulation of an executable query $\hat{q}_b$ in the native language of each $b \in \mathcal{S}$, and the consolidation of the executor outputs $\{\texttt{Exec}(b, \hat{q}_b)\}_{b \in \mathcal{S}}$ into a single evidence set relevant to $q$. This formulation has clear strengths. In particular, since each source is engaged through its own native language, the structural operators it exposes (such as joins, traversals, property paths) are preserved rather than approximated by similarity in a shared space. Also, keeping each source on its own terms makes adding a new source a matter of registration rather than infrastructure rebuilding, and lets the framework draw on any of its registered sources for a single question (whether the answer comes back as a passage, a tuple, a triple, or a path) without committing to one backend up front.

\subsection{Source Selection}

We now turn to the first operation in OmniRetrieval, identifying a subset of sources to engage for a question.

\paragraph{Challenges in Source Selection}
The set of registered sources $\mathcal{B}$ can be large and is open-ended, since new sources are added by registration alone, and the structural contexts $\{c_b\}_{b \in \mathcal{B}}$ that distinguish one source from another are heterogeneous in form (a schema lists tables and columns, an ontology declares classes and predicates, and a corpus descriptor characterizes the topics and style of its documents). One straightforward approach to operationalize selection is to embed each $c_b$ and the query $q$ into a shared vector space and rank sources by similarity, following standard single-corpus retrieval practice. However, this approach is restrictive, since the descriptors are not uniform in form so a single encoder cannot represent them without lossy projection, and the decision of whether $b$ can answer $q$ often hinges on the actual contents of $c_b$ (such as a table name in a relational schema, or a relationship type in a property graph), which a similarity score alone cannot capture.

\paragraph{Long-Context Source Selection}
To sidestep this issue, inspired by recent works showing that long-context LLMs can retrieve and reason directly over textual inputs at the scale of entire corpora~\citep{LOFT, ToTAL}, we propose to read the full catalog of source descriptors jointly with $q$ and identify the sources to engage. Yet, in contrast to such prior works that leverage this in-context capability to homogeneous corpora, our approach involves heterogeneous knowledge sources whose descriptors each take a different form. Formally, a long-context LLM takes the query $q$ and the structural descriptors $\{c_b\}_{b \in \mathcal{B}}$ of all registered sources (schemas, ontologies, and corpus summaries; see Appendix~\ref{sec:appendix_context} for examples) as input, and returns a ranked subset of $\mathcal{B}$:
\begin{equation}
\mathcal{S} = \texttt{LLM}_{\texttt{select}}(q,\, \{c_b\}_{b \in \mathcal{B}};\, k) \subseteq \mathcal{B},
\end{equation}
where the LLM returns at most $k$ sources ordered by relevance to $q$. Since $c_b$ here is the same structural context each source already publishes under the formulation, it could be used as-is and grows by simply appending a new descriptor. More interestingly, the operator returns a short list of candidates, which enables accommodating the queries that require multiple sources as well as the queries whose target source is ambiguous, with the final decision then deferred to the evidence selection stage, where it can rest on the retrieved evidence.

\subsection{Query Formulation}

Given $\mathcal{S}$, we now formulate an executable query $\hat{q}_b$ in the native language of each source $b \in \mathcal{S}$.

\paragraph{Challenges in Query Formulation}
The knowledge sources in $\mathcal{S}$ each speak their own native query language, each shaped by the structure of the data it queries: SQL expresses joins and set operations over normalized relational tables, SPARQL matches triple patterns over an RDF graph, Cypher traverses paths over the labeled nodes and relationships of a property graph, and free-form text drives similarity-based retrieval over a corpus. Beyond the difference in languages, an executable query for $b$ should also refer to the elements that $b$ actually exposes, such as its specific tables and columns in a relational schema, the predicates declared in an RDF ontology, the relationship types of a property graph, or the topical scope and style of the corpus. The proposed framework should therefore produce, for every source $b \in \mathcal{S}$, a query that is both valid in its native query language and grounded in its structural context $c_b$, across distinct backend types.

\paragraph{Per-Source Native Query Generation}
To address this, for each $b \in \mathcal{S}$ the framework translates the question $q$ into a native query $\hat{q}_b$ in the language of $b$, conditioned on the structural context $c_b$:
\begin{equation}
\hat{q}_b = \texttt{Generate}_b(q,\, c_b) \quad \text{for each } b \in \mathcal{S}.
\end{equation}
Here, we instantiate $\texttt{Generate}_b$ as $\texttt{LLM}(\mathcal{T}_b(q,\, c_b))$, with $\texttt{LLM}$ as a single LLM shared across all sources and $\mathcal{T}_b$ a per-source prompt template that incorporates $q$, $c_b$, and an instruction identifying the native query language of $b$. In particular, for SQL, SPARQL, and Cypher, the LLM emits the executable native query directly; for an unstructured corpus, the retriever accepts free-form text, so $q$ itself can serve as the retriever query, and the LLM could optionally be used to optimize $q$ to improve retrieval. Additionally, the LLM-based realization above is one of several possible instantiations of $\texttt{Generate}_b$, and any method that maps $q$ and $c_b$ to a valid native query for $b$ would fit the framework.

\subsection{Cross-Source Evidence Selection}

From a collection of executor outputs through $\texttt{Exec}(b, \cdot)$, we now select the final evidence set $\mathcal{E}$ relevant to $q$.

\paragraph{Role of Selection}
Our task is retrieval, whose objective is to return what is relevant to the question and filter out what is not; however, the executor outputs do not yet meet that goal: running each $\hat{q}_b$ through $\texttt{Exec}(b, \cdot)$ produces an individual retrieval result for each source $b \in \mathcal{S}$, with $\mathcal{S}$ itself possibly spanning multiple sources because the question requires them or because the source-selection step deferred an ambiguous routing decision here. In addition, these per-source results are heterogeneous in both form (rows from SQL, triples from RDF, paths from property graphs, passages from an unstructured corpus) and size, since each $\texttt{Exec}(b, \hat{q}_b)$ returns results at the granularity of the native query, ranging from many items down to a single value such as an entity or an aggregate, only some of which are typically relevant to $q$. The role of this step is therefore to pick, from $\{\texttt{Exec}(b, \hat{q}_b)\}_{b \in \mathcal{S}}$, the subset relevant to $q$, completing the retrieval task by filtering out what is not.

\paragraph{Cross-Source Evidence Selection}
Formally, the OmniRetrieval framework implements this as an operator $\texttt{Select}$ that takes $q$ and the executor outputs, and then returns the relevant subset $\mathcal{E}$:
\begin{equation}
\mathcal{E} = \texttt{Select}(q,\, \{\texttt{Exec}(b, \hat{q}_b)\}_{b \in \mathcal{S}}).
\end{equation}
Here, we instantiate $\texttt{Select}$ as $\texttt{LLM}(\mathcal{T}_{\text{sel}}(\cdot))$, with $\mathcal{T}_{\text{sel}}$ a prompt template that verbalizes each executor output in its native form (rows for SQL, triples for RDF, paths for property graphs, and passages for an unstructured corpus), and asks the model to identify the outputs relevant to $q$. It is worth noting that, although native query languages are essential at the query stage to express structural operators (joins, traversals, paths), verbalizing the executor outputs here does not undercut that choice: by this point $\texttt{Exec}$ has already done the structural work via those operators, providing results that can be read as text.

\section{Experimental Setup}

\subsection{Datasets and Knowledge Bases}

We evaluate OmniRetrieval on a benchmark compiled from 13 datasets that, in combination, span all four native backends, and that together provide a pool of 309 distinct knowledge bases.

\vspace{-0.075in}
\paragraph{Document Search}
For document retrieval over unstructured corpora, whose task is to identify documents that are most relevant to a natural-language query, we use seven datasets of various domains from the BEIR benchmark~\citep{BEIR}: NFCorpus (medical)~\citep{NFCorpus}, SciFact (scientific claim verification)~\citep{SciFact}, FiQA~\citep{FiQA} (financial question answering), MS MARCO (web passages)~\citep{MSMARCO}, FEVER (Wikipedia fact verification)~\citep{FEVER}, Natural Questions (short-answer question answering)~\citep{NQ}, and HotpotQA~\citep{HotpotQA} (multi-hop question answering). Each document collection itself serves as a knowledge base.

\vspace{-0.075in}
\paragraph{Relational Databases}
For text-to-SQL, the task of translating a natural-language question into a SQL query over a relational database, we use Spider~\citep{Spider} and BIRD~\citep{Bird}: Spider brings 206 databases across diverse domains, and BIRD brings a further 80 databases from real-world applications, yielding 286 knowledge bases in total, with each provided as a SQLite database against which the SQL is executed.

\vspace{-0.075in}
\paragraph{RDF Knowledge Graphs}
For text-to-SPARQL, the task of translating a natural-language question into an executable SPARQL query over an RDF knowledge graph (a structured store of subject-predicate-object triples), we use the following three datasets: SimpleQuestions~\citep{SimpleQuestions} for single-triple factoid questions, QALD-10~\citep{QALD10} for hand-curated questions covering factoid and aggregation queries, and LC-QuAD 2.0~\citep{LCQuAD2} for large-scale, compositional questions. As the largest publicly-queryable RDF knowledge graph and the standard target of modern SPARQL benchmarks, Wikidata is the single knowledge base in this backend, against whose public SPARQL endpoint\footnote{\url{https://query.wikidata.org/sparql}} the query is executed.

\vspace{-0.075in}
\paragraph{Labeled Property Graphs}
For text-to-Cypher, the task of translating a natural-language question into a Cypher query over a property graph (whose nodes and edges carry typed labels along with key-value properties dictated by the graph-specific data model), we use Text2Cypher~\citep{Text2Cypher}, which spans 15 graphs from the Neo4j collection, covering various domains (such as movie recommendations, company structures, social networks, and financial investigations), with the generated query executed against the Neo4j endpoint\footnote{\url{neo4j+s://demo.neo4jlabs.com:7687}}.

For each dataset, we sample 300 questions for evaluation. The structural context $c_b$ is a topical descriptor for document collections and a schema for each structured backend (such as relational databases, RDF knowledge graphs, and labeled property graphs); example queries and the verbatim form of each $c_b$ are in Appendix~\ref{sec:appendix_context}.

\subsection{Methods}

We compare OmniRetrieval against three groups of baselines, while holding the backbone model and per-backend execution engines fixed so that any difference traces to how each method engages the pool of sources.

\vspace{-0.025in}
\paragraph{Single-Backend Baselines}
We first include four baselines that pin the pipeline to a single retrieval paradigm and operate on every query irrespective of the underlying knowledge. Specifically, \textbf{Document Search} answers a question through retrieval over an unstructured corpus; \textbf{Text-to-SQL} formulates a SQL query against a relational database; \textbf{Text-to-SPARQL} formulates a SPARQL query against an RDF knowledge graph; and \textbf{Text-to-Cypher} formulates a Cypher query against a labeled property graph. For a fair comparison with our framework, the knowledge base within each paradigm is selected per query in the same manner.

\vspace{-0.025in}
\paragraph{KB Routing}
We further include a baseline that, unlike the four above, lets the model choose any backend per query: it reads the catalog of source descriptors jointly with the question and routes to a single knowledge base, after which query formulation and execution proceed on that source.

\vspace{-0.025in}
\paragraph{OmniRetrieval}
This is the proposed framework, which engages multiple candidate sources: given a question, it reads the source catalog and returns a short list of candidates (e.g., 3), formulates a native query for each, executes them, and consolidates the results through cross-source evidence selection.

\vspace{-0.025in}
\paragraph{Oracle}
As a non-comparable upper bound, this method achieves perfect source selection, using the gold knowledge base annotated in each sample and leaving only query formulation and execution.

\vspace{-0.025in}
\paragraph{Unified-Representation}
A direct comparison against methods that collapse heterogeneous sources into a unified representation~\citep{UniK, UDT, DiFaR} is not realizable, since materializing such a representation is already infeasible at our benchmark scale, itself a small slice of real-world deployments. For example, while Wikipedia underlies several of our corpora at 7 million passages, Wikidata holds well over 15 billion triples\footnote{\url{https://www.wikidata.org/}}, several orders of magnitude beyond typical dense indexes. The same gap recurs for labeled property graphs, where paths (the natural retrieval unit) grow exponentially with hop length and three-hop paths on one graph in our pool already reach tens of billions; and for relational databases, where one database in our pool holds over 70 million rows while row-level encoding further discards joins and set operations SQL is meant to express. Nonetheless, we report these methods under a feasibility-constrained setup in Table~\ref{tab:constrained_baseline}.

\subsection{Evaluation Metrics}

We utilize three metrics covering source selection, retrieval quality, and a soft, judge-based assessment, all macro-averaged across the four native retrieval paradigms so that each contributes equally.

\vspace{-0.025in}
\paragraph{Source Selection Accuracy}
It measures how often a method selects and includes both the correct backend and knowledge base for each question.

\vspace{-0.025in}
\paragraph{Retrieval Accuracy}
It evaluates the quality of retrieved results: \textbf{NDCG@10} for document search (how well the retrieved ranking matches the gold relevance annotations), and \textbf{Execution Match} for SQL, SPARQL, and Cypher (whether the executed result set matches that of the gold query).

\vspace{-0.025in}
\paragraph{LLM-as-a-Judge}
The metrics above are strict: they require the predicted output to match the gold reference exactly, penalizing both the surface-form differences (such as a passage versus a table row) and the selection to a legitimately alternative source. To complement them with a softer signal that tolerates these surface differences, we use an \textbf{LLM-as-a-Judge}~\citep{GEval} with GPT-5.4-mini~\citep{GPT-54}: the judge sees the question, the prediction, and the gold annotation, and credits the prediction whenever the predicted output is semantically equivalent to gold or faithfully realizes the question against an alternative knowledge base.

\subsection{Implementation Details}

For each comparison, we instantiate all methods with the same backbone, which spans GPT-5.4~\citep{GPT-54}, Gemini-3.1 (Pro)~\citep{Gemini31Pro}, Sonnet-4.6~\citep{Sonnet46}, Qwen-3.5 (27B)~\citep{Qwen3.5}, and Gemma-4 (31B)~\citep{Gemma4}, where closed-source backbones are served through their APIs, while open-source ones are served locally with vLLM~\citep{vLLM}. For document retrieval, we use \texttt{all-MiniLM-L6-v2}~\citep{SBERT} as the shared encoder, but, instead of embedding the question directly, it is first rewritten into a hypothetical passage and then embedded. For text-to-SPARQL, the entity-linking step follows the procedure from~\citet{ToG}, used to build the context $c_b$. Further implementation details are in Appendix~\ref{sec:appendix_implementation}, and the prompts are in Appendix~\ref{sec:appendix_prompts}.

\providecolor{mygray}{HTML}{F2F2F2}
\providecolor{citeblue}{HTML}{1668b0}

\begin{table*}[t]
    \centering
    \caption{
    Main results, with each metric macro-averaged across the four retrieval paradigms. Best results among the comparable methods are \textbf{bolded}; second best are \underline{underlined}. \textit{Oracle} is a upper bound with perfect source selection.
    }
    \label{tab:main}
    \vspace{-0.075in}
    \small
    \resizebox{\textwidth}{!}{
    \setlength{\tabcolsep}{6pt}
    \renewcommand{\arraystretch}{1.0}
    \begin{tabular}{l c c c c c @{\hspace{1.0em}} c}
        \toprule
        \textbf{Method} & \textbf{GPT-5.4} & \textbf{Gemini-3.1 (Pro)} & \textbf{Sonnet-4.6} & \textbf{Qwen-3.5 (27B)} & \textbf{Gemma-4 (31B)} & \textbf{Average} \\

        \midrule

        \rowcolor{mygray}
        \multicolumn{7}{c}{\rule{0pt}{2.0ex}\textbf{\textit{Source Selection Accuracy}}} \\
        \addlinespace[1pt]

        \midrule

        Document Search & 21.49 & 22.61 & 21.90 & 21.49 & 21.42 & 21.78 \\
        Text-to-SQL     & 14.75 & 16.71 & 16.00 & 12.62 & 13.58 & 14.73 \\
        Text-to-SPARQL  & 24.92 & 25.00 & 24.81 & 24.94 & 24.56 & 24.84 \\
        Text-to-Cypher  & 19.92 & 20.42 & 20.83 & 20.42 & 19.08 & 20.13 \\
        KB Routing      & \underline{64.88} & \underline{68.21} & \underline{60.40} & \underline{54.83} & \underline{59.93} & \underline{61.65} \\
        \rowcolor{citeblue!10}
        \textbf{OmniRetrieval (Ours)} & \textbf{68.58} & \textbf{73.30} & \textbf{66.47} & \textbf{57.81} & \textbf{62.40} & \textbf{65.71} \\
        \cdashline{1-7}\addlinespace[0.4ex]
        \textit{Oracle} & \textit{100.00} & \textit{100.00} & \textit{100.00} & \textit{100.00} & \textit{100.00} & \textit{100.00} \\

        \midrule
        \addlinespace[2pt]

        \rowcolor{mygray}
        \multicolumn{7}{c}{\rule{0pt}{2.0ex}\textbf{\textit{Retrieval Accuracy}}} \\
        \addlinespace[1pt]

        \midrule

        Document Search & 13.42 & 14.94 & 13.69 & 13.23 & 13.16 & 13.69 \\
        Text-to-SQL     & 13.51 & 16.75 & 15.46 & 12.78 & 13.92 & 14.48 \\
        Text-to-SPARQL  & 18.26 & 20.71 & 15.60 & 16.65 & 17.93 & 17.83 \\
        Text-to-Cypher  & 18.06 & 17.17 & 18.68 & 18.58 & 17.14 & 17.93 \\
        KB Routing      & \underline{42.07} & \underline{46.83} & \underline{38.59} & \underline{34.28} & \underline{38.13} & \underline{39.98} \\
        \rowcolor{citeblue!10}
        \textbf{OmniRetrieval (Ours)} & \textbf{46.62} & \textbf{52.69} & \textbf{43.07} & \textbf{38.34} & \textbf{40.97} & \textbf{44.34} \\
        \cdashline{1-7}\addlinespace[0.4ex]
        \textit{Oracle} & \textit{62.47} & \textit{65.56} & \textit{60.27} & \textit{60.11} & \textit{60.85} & \textit{61.85} \\

        \midrule
        \addlinespace[2pt]

        \rowcolor{mygray}
        \multicolumn{7}{c}{\rule{0pt}{2.0ex}\textbf{\textit{LLM-as-a-Judge}}} \\
        \addlinespace[1pt]

        \midrule

        Document Search & 39.93 & 40.92 & 40.82 & 36.31 & 39.47 & 39.49 \\
        Text-to-SQL     & 25.61 & 25.86 & 29.63 & 22.00 & 25.16 & 25.65 \\
        Text-to-SPARQL  & 28.89 & 30.29 & 31.06 & 23.99 & 25.70 & 27.99 \\
        Text-to-Cypher  & 33.09 & 24.36 & 34.85 & 24.42 & 25.19 & 28.38 \\
        KB Routing      & \underline{60.26} & \underline{63.67} & \underline{61.93} & \underline{50.37} & \underline{53.71} & \underline{57.99} \\
        \rowcolor{citeblue!10}
        \textbf{OmniRetrieval (Ours)} & \textbf{69.72} & \textbf{71.10} & \textbf{68.62} & \textbf{60.83} & \textbf{59.13} & \textbf{65.88} \\
        \cdashline{1-7}\addlinespace[0.4ex]
        \textit{Oracle} & \textit{75.20} & \textit{76.50} & \textit{76.29} & \textit{71.93} & \textit{72.81} & \textit{74.55} \\

        \bottomrule
    \end{tabular}
    }
\end{table*}

\section{Experimental Results and Analyses}
\label{sec:results}

\paragraph{Main Results}
We report the main results in Table~\ref{tab:main}, where OmniRetrieval consistently outperforms all baselines across the five backbones. The four single-backend baselines, each pinned to one paradigm, perform poorly since three of the four query types lie outside their reach. KB Routing lifts this restriction by routing to one source per query, but its up-front commitment leaves no fallback when the chosen source is wrong. OmniRetrieval instead engages multiple candidates and consolidates them via cross-source evidence selection, yielding consistent gains over KB Routing. Also, the gap to the Oracle upper bound narrows from selection to judge ($34.27 \to 17.51 \to 8.67$ pts), suggesting that evidence selection often recovers a semantically equivalent answer from an alternative source even when source selection misses.

\begin{figure*}[t!]
    \centering
    \begin{minipage}[t]{0.70\linewidth}
        \centering
        \includegraphics[width=0.975\linewidth]{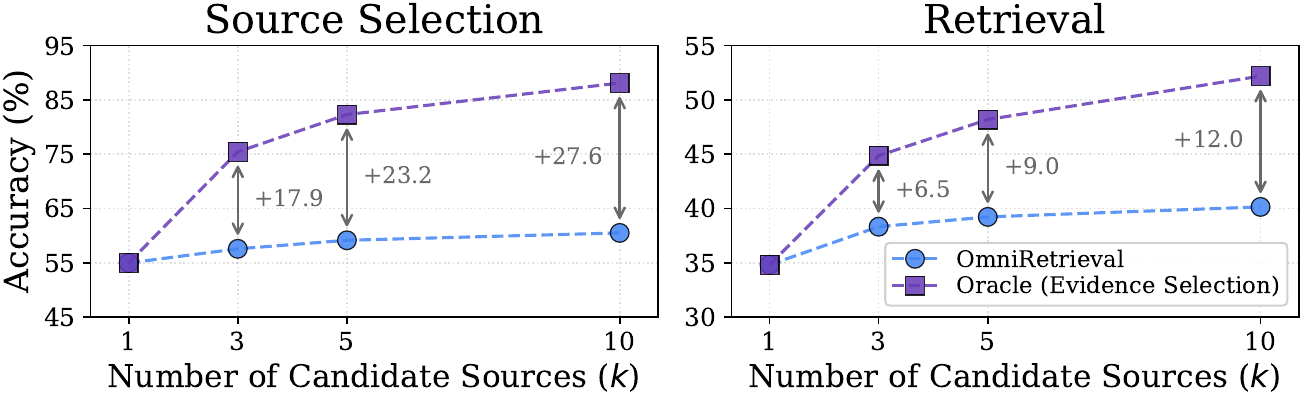}
        \vspace{-0.1in}
        \captionof{figure}{Effect of the candidate list size $k$ in source selection on Source Selection and Retrieval accuracy. \textbf{Oracle (Evidence Selection)} replaces the LLM-based evidence selection with gold-source selection within the top-$k$ candidates.}
        \label{fig:topk_sweep}
    \end{minipage}
    \hfill
    \begin{minipage}[t]{0.28\linewidth}
        \centering
        \includegraphics[width=0.975\linewidth]{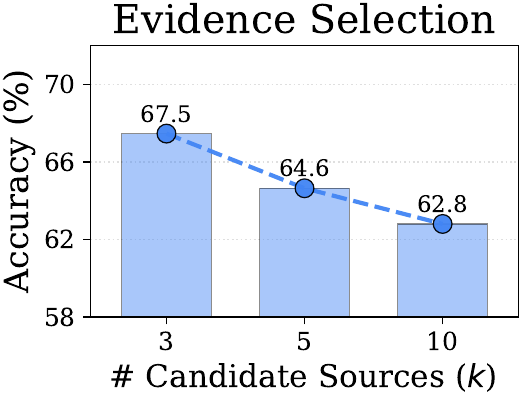}
        \vspace{-0.1in}
        \captionof{figure}{Evidence-selection accuracy on multi-candidate questions with the gold in the top-$k$.}
        \label{fig:selector_analysis}
    \end{minipage}
\end{figure*}

\begin{figure*}[t!]
    \centering
    \begin{minipage}[t]{0.70\linewidth}
        \centering
        \includegraphics[width=0.975\linewidth]{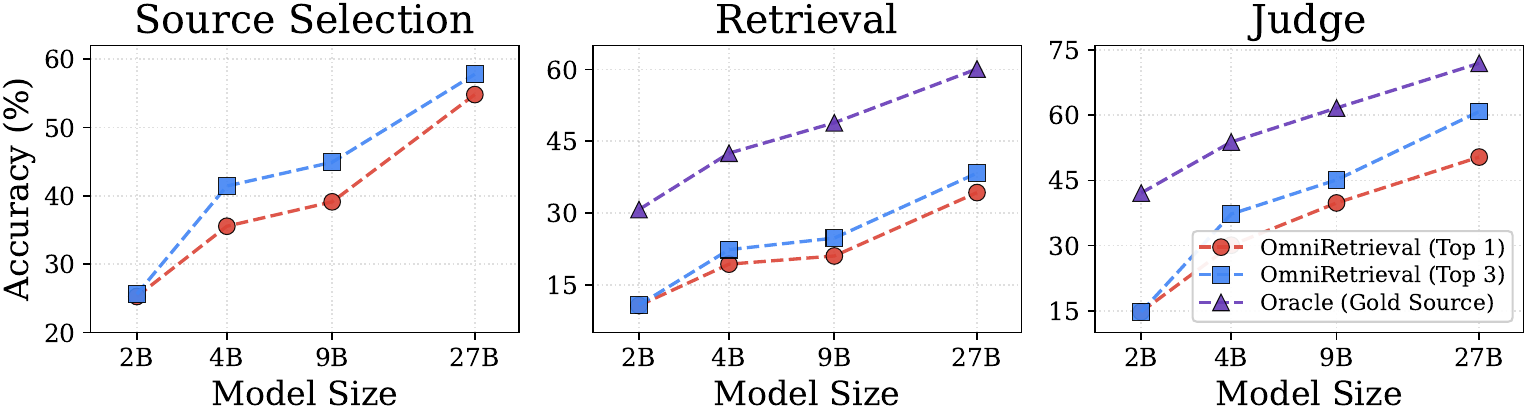}
        \vspace{-0.1in}
        \captionof{figure}{Effect of backbone scale (Qwen-3.5, 2B to 27B). \textbf{Oracle (Gold Source)} uses the gold source in place of LLM-based source selection, and is omitted from Source Selection (where it is trivially $100\%$ by construction).}
        \label{fig:qwen_scaling}
    \end{minipage}
    \hfill
    \begin{minipage}[t]{0.28\linewidth}
        \centering
        \includegraphics[width=0.975\linewidth]{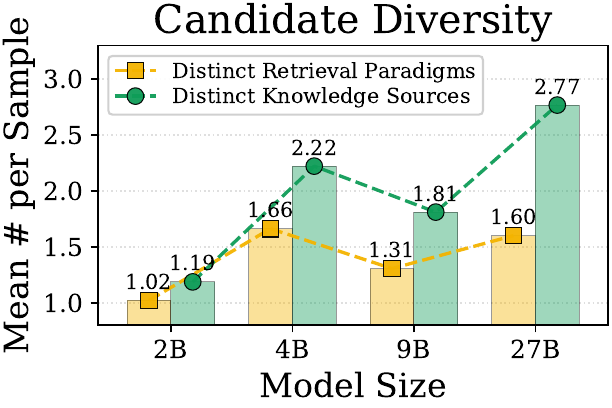}
        \vspace{-0.1in}
        \captionof{figure}{Candidate diversity: distinct retrieval paradigms and knowledge sources per sample.}
        \label{fig:qwen_diversity}
    \end{minipage}
\end{figure*}

\paragraph{Analysis on Source Candidate Size}
Recall that the source-selection step of OmniRetrieval returns a short list of $k$ candidate sources, which is then fed into per-source query formulation and cross-source evidence selection. To see its effect, we sweep $k \in \{1, 3, 5, 10\}$ with Qwen-3.5 (27B) and additionally compare against an \textbf{Oracle (Evidence Selection)} variant that replaces the LLM-based selector with gold-source selection within the $k$. As shown in Figure~\ref{fig:topk_sweep}, OmniRetrieval improves monotonically with $k$, yet the oracle improves much faster, widening the gap to the oracle as $k$ grows. This observation traces to the selector itself: its multi-candidate 1-of-$k$ accuracy drops from $67.5\%$ at $k{=}3$ to $62.8\%$ at $k{=}10$ (Figure~\ref{fig:selector_analysis}). Thus, combined with the linear cost of additional candidates, this points to the initial evidence selection as the more impactful lever.

\paragraph{Analysis on Backbone Scale}
Building on the analysis above, we next vary the backbone scale by sweeping Qwen-3.5 from $2$B to $27$B, comparing OmniRetrieval (Top 1, one candidate per question) with OmniRetrieval (Top 3, our default). As shown in Figure~\ref{fig:qwen_scaling}, the two are essentially tied at 2B and Top 3 takes a clear lead at 27B, a gap that traces to candidate diversity (Figure~\ref{fig:qwen_diversity}): at 2B the source-selection step collapses to a single paradigm, while beyond 4B it produces meaningfully different candidates across both paradigms and sources\footnote{Qwen-3.5 9B exhibits a bias toward Document Search at its top candidate, pulling its candidates into the same retrieval paradigm; this surfaces as the diversity dip in Figure~\ref{fig:qwen_diversity}.}. Yet the gap to Oracle (Gold Source) ceiling remains, the largest on Source Selection, underscoring source selection as the most consequential step in the pipeline. This aligns with the design rationale of OmniRetrieval: broad exploration at source-selection, with the final commitment deferred to the evidence-selection.

\begin{figure*}[t!]
    \centering
    \begin{minipage}[t]{0.36\linewidth}
        \vspace{0pt}
        \centering
        \includegraphics[width=0.98\linewidth]{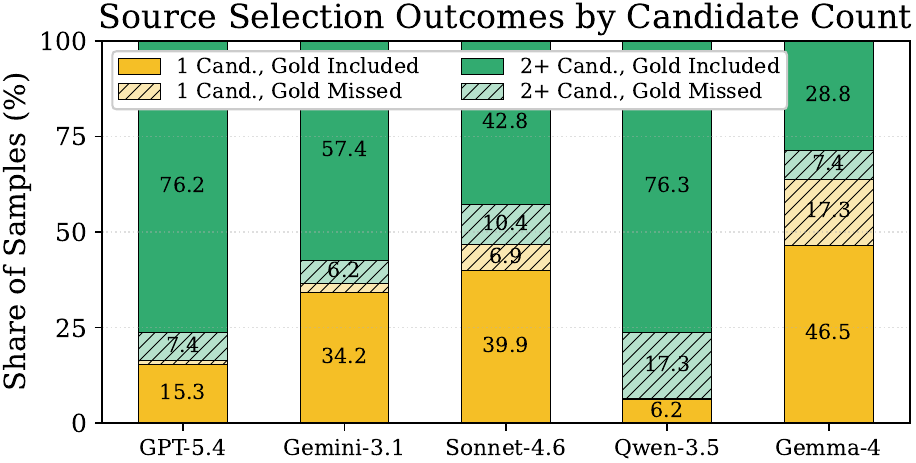}
        \vspace{-0.11in}
        \captionof{figure}{Source-selection behavior on 1 and 2+ candidate regimes. Solid segments mark success (gold present in the candidate list); hatched ones mark failure.}
        \label{fig:cross_backbone_selector}
    \end{minipage}
    \hfill
    \begin{minipage}[t]{0.315\linewidth}
        \vspace{0pt}
        \centering
        \captionof{table}{Evidence-selection accuracy on multi-candidate questions containing the gold, where \textit{Random} is the per-backbone uniform-pick baseline.}
        \label{tab:evidence_selection}
        \vspace{-0.05in}
        \resizebox{\linewidth}{!}{%
        \setlength{\tabcolsep}{2.5pt}
        \renewcommand{\arraystretch}{1.25}
        \begin{tabular}{l c c c}
            \toprule
            \textbf{Backbone} & \textbf{Acc. (\%)} & \textbf{Rand. (\%)} & \textbf{$\Delta$ (pp)} \\
            \midrule
            GPT-5.4    & 72.81 & 38.31 & $+34.51$ \\
            Gemini-3.1 & 75.29 & 43.99 & $+31.30$ \\
            Sonnet-4.6 & 70.44 & 41.47 & $+28.98$ \\
            Qwen-3.5   & 67.91 & 35.55 & $+32.36$ \\
            Gemma-4    & 74.33 & 47.73 & $+26.60$ \\
            \bottomrule
        \end{tabular}%
        }
    \end{minipage}
    \hfill
    \begin{minipage}[t]{0.30\linewidth}
        \vspace{0pt}
        \centering
        \includegraphics[width=0.95\linewidth]{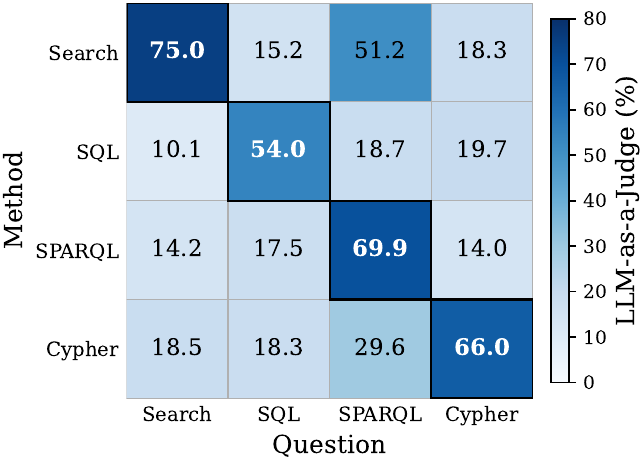}
        \vspace{-0.11in}
        \captionof{figure}{LLM-Judge accuracy on GPT-5.4 (rows: method paradigm; columns: question paradigm).}
        \label{fig:paradigm_coverage}
    \end{minipage}
\end{figure*}

\paragraph{Analysis on Cross-Source Evidence Selection}
Building on the prior deferral argument, we examine how reliably evidence selection commits to the gold candidates. The source-selection step lands in one of two regimes per question, either committing to a single candidate or emitting two or more, and the exploration versus commitment trade-off varies widely across backbones (Figure~\ref{fig:cross_backbone_selector}). Yet two patterns hold consistently. First, the gold source is included in the candidate list at a high rate at every backbone. Second, once included, the evidence-selection step picks it at a high rate as well, far above the per-backbone random baseline (Table~\ref{tab:evidence_selection}). Broad upstream exploration thus translates into accurate final commitments at the step that follows.

\paragraph{Analysis on Cross-Paradigm Coverage}
To see how much of the question space each paradigm can cover, Figure~\ref{fig:paradigm_coverage} reports a matrix of LLM-as-a-Judge accuracy on GPT-5.4, with rows as the method paradigm and columns as the question paradigm. From this, we find that Document Search has by far the widest cross-paradigm coverage, with an off-diagonal mean of $28.2\%$ against $15.2$ to $22.1\%$ for the three structured paradigms. The advantage is largely driven by SPARQL questions, where the Wikipedia-derived corpora exposed under Search overlap with the factual content of Wikidata.

\providecolor{mygray}{HTML}{F2F2F2}
\providecolor{citeblue}{HTML}{1668b0}

\begin{wraptable}{r}{0.45\textwidth}
    \vspace{-0.20in}
    \centering
    \caption{
    Results for the unified-representation methods~\citep{UniK, UDT} with the constrained setup on GPT-5.4, marked $^\dagger$ as non-comparable.
    }
    \label{tab:constrained_baseline}
    \vspace{-0.075in}
    \small
    \resizebox{\linewidth}{!}{
    \setlength{\tabcolsep}{3pt}
    \renewcommand{\arraystretch}{1.1}
    \begin{tabular}{l c c c}
        \toprule
        \textbf{Method} & \textbf{Source Sel.} & \textbf{Retrieval} & \textbf{Judge} \\

        \midrule

        Document Search                              & 21.49 & 13.42 & 39.93 \\
        Unified Representation$^\dagger$             & 31.00 & 23.00 & 45.00 \\
        KB Routing                                   & \underline{64.88} & \underline{42.07} & \underline{60.26} \\
        \rowcolor{citeblue!10}
        \textbf{OmniRetrieval (Ours)}                & \textbf{68.58} & \textbf{46.62} & \textbf{69.72} \\

        \bottomrule
    \end{tabular}
    }
    \vspace{-0.1in}
\end{wraptable}

\paragraph{Analysis on Native vs Unified Retrieval}
To examine how shared-representation methods would perform once the pool is shrunk to a scale they can index (via sub-sampling), we build a constrained pool that retains only the gold-touched triples and edges in SPARQL and Cypher (with a same-count random distractor), includes all SQL tables, and keeps the document corpus at full scale. All items are verbalized and embedded into a single space with \texttt{all-MiniLM-L6-v2}, and the top-3 retrievals, matched to the candidate budget of OmniRetrieval, serve as the final results. As shown in Table~\ref{tab:constrained_baseline}, the unified method surpasses the four single-backend baselines through occasional cross-paradigm coverage, but stays far below KB Routing and OmniRetrieval. The gap persists even though the constrained setting hands this unified method an advantage no other method receives, pointing to a fundamental limit: atomic-unit retrieval cannot capture the structural composition (e.g., joins, traversals, multi-hop chains) that native queries express.

\section{Related Work}

\paragraph{Retrieval over Heterogeneous Sources}
Classical retrieval has long been organized around a single corpus and representation, from lexical retrievers that rank passages by term overlap~\citep{BM25, tfidf} to dense retrievers that project queries and documents into an embedding space~\citep{DPR, ANCE}, with multi-modal extensions adding specific encoders for images or video~\citep{CLIP, VideoRAG}. To lift this single-source restriction, one line of work collapses heterogeneous sources, such as text passages, knowledge graph facts, and tabular records, into a shared representation so that a single retriever can rank items across them~\citep{UniK, UDT, DiFaR}. In the meantime, other efforts cover either structured or unstructured sources but not both: query-type routers are confined to unstructured corpora and rely on embedding similarity within each~\citep{UniversalRAG}, while hand-crafted, per-type interface functions are confined to structured sources and further presuppose a single target source fixed at input rather than chosen per question, with native query generation only for relational databases~\citep{StructGPT}. Yet none of these approaches generate queries uniformly in the native language of each backend: structure is flattened into a common representation, surfaced only through embedding similarity, or fragmented across per-type interface functions with predefined extraction primitives. In contrast, OmniRetrieval covers both structured and unstructured backends, formulating a query for each in its native language and thereby preserving the symbolic operators each source exposes.

\paragraph{Native-Language Query Generation}
A substantial body of work studies how to translate a natural-language query into the native query language of a single backend. Text-to-SQL has matured around schema-grounded generation over relational databases, with widely used benchmarks~\citep{Spider, Bird} and capable LLM-based generators that perform schema linking and SQL synthesis~\citep{Text2SQL-LLM1, Text2SQL-LLM2}. Parallel lines of work address text-to-SPARQL for RDF stores~\citep{SP-LLM1, SP-LLM2, Text2SPARQL} and text-to-Cypher for labeled property graphs~\citep{GPT4Graph, Text2Cypher}, each respecting the syntax and semantics of the target language. OmniRetrieval inherits this per-backend capability and lifts it into a multi-backend setting, where any existing generator can be plugged in as the query-formulation module, while introducing challenges intrinsic to this regime: joint source selection, query formulation across multiple native languages within one model, and cross-form evidence consolidation.

\paragraph{LLMs as Tool-Use Agents}
A broader paradigm casts LLMs as agents that interact with external tools via API calls to extend their capabilities beyond parametric memory~\citep{ReAct, Toolformer, ToolLLM, Gorilla}. OmniRetrieval shares this dispatch pattern but is, to our knowledge, the first framework to unify retrieval across heterogeneous structured and unstructured backends through their native query languages, and it differs from generic tool use in two concrete ways. First, each invocation is a program synthesis grounded in a backend schema that can span hundreds of tables or thousands of relations, well beyond the small fixed function signatures typical of tool use. Second, results are structurally heterogeneous (passages, table rows, KG triples, graph paths) and require cross-form consolidation into a unified evidence set, a step that does not arise in tool use where each output is consumed independently.

\section{Conclusion}

In this work, we presented OmniRetrieval, a framework for retrieval over structurally heterogeneous knowledge sources that, given a natural-language question, engages each relevant source through its own native query language and consolidates the executor outputs via a cross-source evidence selection step, rather than collapsing the sources into a shared representation. Evaluation on a benchmark spanning 13 datasets and 309 knowledge bases over unstructured corpora, relational databases, RDF graphs, and labeled property graphs shows that OmniRetrieval consistently outperforms relevant baselines. Our analyses further indicate that broad exploration at the source-selection step, with the final commitment deferred to a selector that rests on retrieved evidence, is what lets OmniRetrieval scale gracefully. These findings position OmniRetrieval as a step toward a general-purpose universal layer, one that preserves the structural affordances that make each source valuable while exposing a single natural-language interface to the user.

\section*{Limitations}

OmniRetrieval is a general framework, and our specific instantiation in this work opens a couple of directions for further exploration. First, while our instantiation of the cross-source evidence selection step already performs reliably, it would be valuable to further strengthen this in future work, for example, through supervised fine-tuning on labeled cross-source selections or reinforcement learning that uses downstream answer quality as the reward signal. Second, our setup uses a single shared LLM across source selection, native query formulation, and evidence selection, and operator-specific specialization is another avenue worth investigating.

\section*{Ethical Considerations}

Our work builds on publicly available datasets and standard LLMs, and we do not foresee ethical concerns beyond those inherited from these underlying components. In particular, since the retrieved evidence and the generated source-native queries draw on the connected knowledge bases and the internalized knowledge within the LLMs, any private, harmful, or biased content present in these resources may carry over to the outputs, and we recommend applying standard safeguards and filtering over both the connected sources and the generated queries to ensure safe and responsible deployment.

\bibliographystyle{plainnat}
\bibliography{paper}

\clearpage
\appendix{
\section{Benchmark Details}
\label{sec:appendix_context}

This section provides additional details on our benchmark, including example questions (Table~\ref{tab:appendix_examples}), the verbatim structural context $c_b$ for each backend, and a note on the use of existing artifacts.

\begin{table*}[t]
\centering
\caption{One example question per dataset, covering the 13 datasets that together span the four native backends.}
\label{tab:appendix_examples}
\vspace{-0.05in}
\small
\resizebox{\textwidth}{!}{
\begin{tabular}{lll}
\toprule
\textbf{Backend} & \textbf{Dataset} & \textbf{Example Question} \\
\midrule
\multirow{7}{*}{Document Search}
  & NFCorpus         & ``Cancer Risk From French Fries'' \\
  & SciFact          & ``Myelin sheaths play a role in action potential propagation.'' \\
  & FiQA             & ``Credit card closed. Effect on credit score?'' \\
  & MS MARCO         & ``where is wetransfer based'' \\
  & FEVER            & ``Sean Connery was cast in The Rock.'' \\
  & Natural Questions & ``who was originally offered the role in pretty woman'' \\
  & HotpotQA         & ``Which software application was originally made for the company that acquired RadioShack in 1963?'' \\
\midrule
\multirow{2}{*}{Relational Databases}
  & Spider           & ``What are all distinct countries where singers above age 20 are from?'' \\
  & BIRD             & ``How many online purchases did Ole Group make in May 2019?'' \\
\midrule
\multirow{3}{*}{RDF Knowledge Graphs}
  & SimpleQuestions  & ``Who discovered 20468 petercook?'' \\
  & QALD-10          & ``Which artists were born on the same date as Rachel Stevens?'' \\
  & LC-QuAD 2.0      & ``What periodical literature does Delta Air Lines use as a mouthpiece?'' \\
\midrule
Labeled Property Graphs
  & Text2Cypher      & ``Which actors have acted in movies directed by the person who directed `Speed Racer'?'' \\
\bottomrule
\end{tabular}
}
\end{table*}

\subsection{Document Search}
\label{sec:appendix_context_search}
For each corpus, $c_b$ is a short topical descriptor comprising a one-line description of the domain, and the typical document style. The example below is the verbatim $c_b$ used for NFCorpus.

\begin{tcolorbox}[colback=gray!5, colframe=gray!50, boxrule=0.5pt, arc=2mm, left=4pt, right=4pt, top=3pt, bottom=3pt]
{\footnotesize
\begin{verbatim}
Corpus: nfcorpus
Description: Medical and nutritional information retrieval
Query type: a consumer-health or nutrition query, often very terse,
  either a one-to-three-word topic stub (a food, condition, or substance)
  or a popular-science headline phrased for a lay audience
Document style: a PubMed-style biomedical research abstract with a formal
  scientific title, written in technical register
\end{verbatim}
}
\end{tcolorbox}

\subsection{Relational Databases}
\label{sec:appendix_context_sql}
For each relational database, $c_b$ is the schema, exposed as the \texttt{CREATE TABLE} declarations of all tables, including column types, primary keys, and foreign keys. The example below is the verbatim $c_b$ for the Spider database \texttt{concert\_singer} (one of four tables shown; the other three are omitted).

\begin{tcolorbox}[colback=gray!5, colframe=gray!50, boxrule=0.5pt, arc=2mm, left=4pt, right=4pt, top=3pt, bottom=3pt]
{\footnotesize
\begin{verbatim}
CREATE TABLE "singer" (
  "Singer_ID" int,
  "Name" text,
  "Country" text,
  "Song_Name" text,
  "Song_release_year" text,
  "Age" int,
  "Is_male" bool,
  PRIMARY KEY ("Singer_ID")
)
[CREATE TABLE statements for concert, singer_in_concert, and stadium omitted]
\end{verbatim}
}
\end{tcolorbox}

\subsection{RDF Knowledge Graphs}
\label{sec:appendix_context_sparql}
For Wikidata, $c_b$ is built per question and combines (i) a fixed preamble of SPARQL prefixes and schematic query templates with (ii) the topic entities mentioned in the question, linked to Wikidata QIDs, and (iii) the top-30 candidate predicates per entity, ranked by semantic similarity between the predicate label and the question. The example below is an abridged $c_b$ for the LC-QuAD 2.0 question shown in Table~\ref{tab:appendix_examples}.

\begin{tcolorbox}[colback=gray!5, colframe=gray!50, boxrule=0.5pt, arc=2mm, left=4pt, right=4pt, top=3pt, bottom=3pt]
{\footnotesize
\begin{verbatim}
Knowledge graph: Wikidata
Prefixes: wd: (entity), wdt: (direct/truthy property), p: (entity -> statement node),
  ps: (statement -> main value), pq: (statement -> qualifier value), rdfs: (for rdfs:label)
Format examples (schematic IDs; use the actual topic entities and relations for the query):
  SELECT ?x WHERE { wd:Qxxx wdt:Pyyy ?x }
  SELECT ?x WHERE { wd:Qxxx wdt:Pyyy ?y . ?y wdt:Pzzz ?x }
  ASK WHERE { wd:Qxxx wdt:Pyyy wd:Qzzz }
  [count, filter, order-by forms omitted]

Topic entities (from the question, already linked to Wikidata QIDs):
- wd:Q188920  (Delta Air Lines)
- wd:Q1002697 (periodical)

Linked relations (candidate properties per topic entity, choose among these):
- wd:Q188920  (Delta Air Lines):
    - P31   (instance of)
    - P229  (IATA airline designator)
    - P2813 (house publication)
    [further candidates omitted]
- wd:Q1002697 (periodical):
    - P31  (instance of)
    [further candidates omitted]
\end{verbatim}
}
\end{tcolorbox}

\subsection{Labeled Property Graphs}
\label{sec:appendix_context_cypher}
For each Text2Cypher graph, $c_b$ is the graph schema, exposed as node labels with typed property keys, relationship types with typed property keys, and the list of edges as triples. The example below is the verbatim $c_b$ for the database \texttt{neo4jlabs\_demo\_db\_movies}.

\begin{tcolorbox}[colback=gray!5, colframe=gray!50, boxrule=0.5pt, arc=2mm, left=4pt, right=4pt, top=3pt, bottom=3pt]
{\footnotesize
\begin{verbatim}
Node properties:
- Movie
  - `title`: STRING Example: "The Matrix"
  - `votes`: INTEGER Min: 1, Max: 5259
  - `tagline`: STRING
    Example: "Welcome to the Real World"
  - `released`: INTEGER Min: 1975, Max: 2012
- Person
  - `born`: INTEGER Min: 1929, Max: 1996
  - `name`: STRING Example: "Keanu Reeves"
Relationship properties:
- ACTED_IN
  - `roles: LIST` Min Size: 1, Max Size: 6
- REVIEWED
  - `summary: STRING` Available options: [...]
  - `rating: INTEGER` Min: 45, Max: 100
The relationships:
(:Person)-[:ACTED_IN]->(:Movie)
(:Person)-[:DIRECTED]->(:Movie)
(:Person)-[:PRODUCED]->(:Movie)
(:Person)-[:WROTE]->(:Movie)
(:Person)-[:FOLLOWS]->(:Person)
(:Person)-[:REVIEWED]->(:Movie)
\end{verbatim}
}
\end{tcolorbox}

\subsection{Use of Existing Artifacts}
\label{sec:appendix_artifacts}
All artifacts used in our benchmark (the 13 datasets and 309 knowledge bases) are existing public resources, and we use each under its respective license and terms. Our use is restricted to research purposes, consistent with the intended use specified by the original releases, and we do not perform additional filtering for personally identifying information or offensive content, deferring to the policies of the upstream datasets.

\section{Implementation Details}
\label{sec:appendix_implementation}
We use a sampling temperature of $0.0$ across all LLM calls (source selection, query formulation, and cross-source evidence selection), which makes the predictions deterministic, so all reported numbers come from a single run per configuration. The maximum number of generated tokens is capped at $1024$. Open-source backbones are served locally on a single NVIDIA H200 GPU.

\section{Per-Paradigm Breakdown}
\label{sec:appendix_per_paradigm}
Table~\ref{tab:per_route_breakdown} reports the per-paradigm decomposition of the three metrics from Table~\ref{tab:main}, broken down by the four native retrieval paradigms (Search, SQL, SPARQL, Cypher) at each of the five backbones. The macro average across the four columns recovers the corresponding column in Table~\ref{tab:main}.

\section{Paradigm Prediction Balance}
\label{sec:appendix_paradigm_balance}
To examine how source selection allocates predictions across the four retrieval paradigms, we report the predicted paradigm distribution in Figure~\ref{fig:paradigm_balance}, under per-paradigm balanced weighting in which each gold paradigm contributes equally to the totals, so the reference becomes a uniform $25\%$ line per paradigm. We show both the top-1 candidate and the full pool of emitted candidates. From this, we observe that the predicted distribution stays within a narrow band of the balanced reference across all five backbone models: SQL ranges over $30$ to $37\%$ at top-1, and Search ranges over $27$ to $37\%$, indicating that source selection remains broadly balanced across paradigms even though the underlying catalog is dominated by relational databases ($\approx 93\%$, 286 of 309 knowledge sources).

\begin{figure*}[t!]
    \centering
    \includegraphics[width=0.95\linewidth]{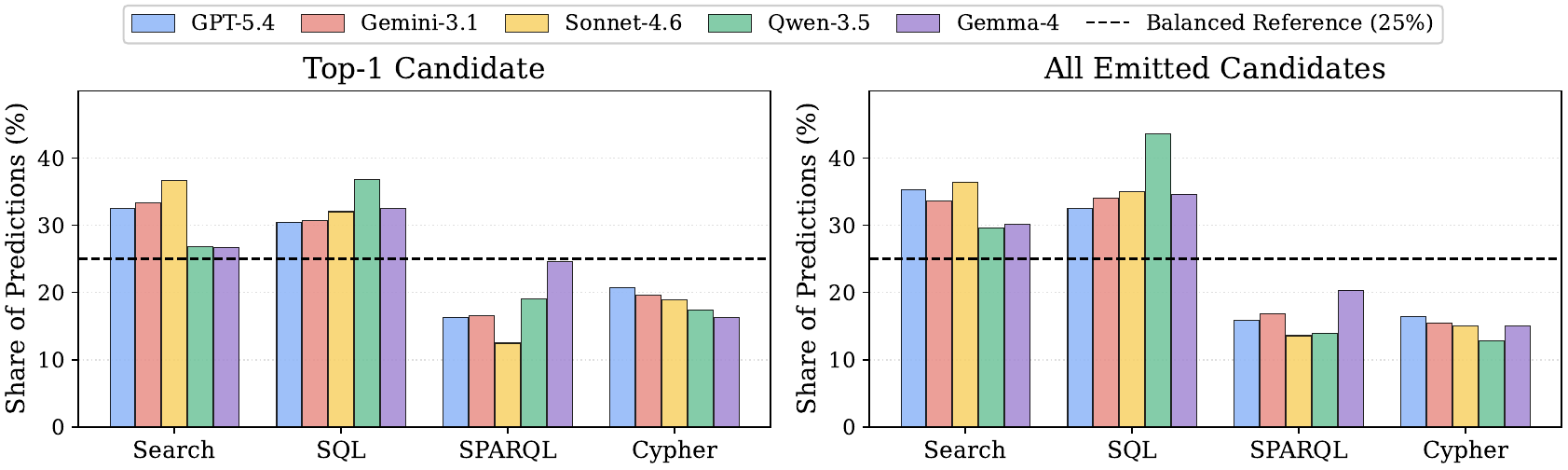}
    \vspace{-0.05in}
    \captionof{figure}{Predicted retrieval paradigm distribution under per-paradigm balanced weighting. \textbf{Left:} top-1 candidate per question. \textbf{Right:} all emitted candidates. The dashed line marks the uniform $25\%$ reference.}
    \label{fig:paradigm_balance}
\end{figure*}

\section{Prompts}
\label{sec:appendix_prompts}
This section lists the prompts used in our framework, covering the source selection, query formulation, and cross-source evidence selection steps, as well as the LLM-as-a-Judge metric.

\vspace{-0.1in}
\paragraph{Source Selection}
Figure~\ref{fig:source_selection_prompt} shows the prompt used for source selection. Given a catalog of available knowledge bases grouped by backend, where each knowledge base is summarized by a short descriptor (such as description, examples, or schema summary), the model returns up to $k$ ranked candidate sources for the question.

\vspace{-0.1in}
\paragraph{Query Formulation}
Figure~\ref{fig:query_formulation_prompt} shows the system prompts for the three native query languages and the shared user template that grounds the query in the backend context $c_b$. For document retrieval, the question is additionally rewritten into a hypothetical passage with the prompt in Figure~\ref{fig:search_rewriting_prompt}, similar to~\citet{HyDE}.

\vspace{-0.1in}
\paragraph{Cross-Source Evidence Selection}
Figure~\ref{fig:evidence_selection_prompt} shows the prompt for picking the best candidates among multiple source-specific retrieval results.

\vspace{-0.1in}
\paragraph{LLM-as-a-Judge}
Figure~\ref{fig:judge_prompt} shows the prompt used by the LLM-as-a-Judge metric, used to assess whether the predicted answer answers the question by direct match or faithful realization against gold and question on an alternative knowledge base.

\providecolor{mygray}{HTML}{F2F2F2}
\providecolor{citeblue}{HTML}{1668b0}

\begin{table*}[t]
    \centering
    \caption{
    Per-paradigm decomposition of Table~\ref{tab:main}, broken down by the four native retrieval paradigms. Each cell is reported as ``$R$ / $E$ / $J$'', where $R$ is Source Selection Accuracy (both the backend and knowledge base correct), $E$ is Retrieval Accuracy (NDCG@10 for Search, Execution Match for the structured backends), and $J$ is LLM-as-a-Judge accuracy.
    }
    \label{tab:per_route_breakdown}
    \vspace{-0.05in}
    \scriptsize
    \resizebox{\textwidth}{!}{
    \setlength{\tabcolsep}{3pt}
    \renewcommand{\arraystretch}{1.1}
    \begin{tabular}{l l c c c c}
        \toprule
        \textbf{Backbone} & \textbf{Method} & \textbf{Search} ($R$ / $E$ / $J$) & \textbf{SQL} ($R$ / $E$ / $J$) & \textbf{SPARQL} ($R$ / $E$ / $J$) & \textbf{Cypher} ($R$ / $E$ / $J$) \\
        \midrule

        \multirow{7}{*}{GPT-5.4}
          & Document Search           & 85.95 / 48.62 / 75.00 &  0.00 /  2.17 / 15.17 &  0.00 /  0.22 / 51.22 &  0.00 /  2.67 / 18.33 \\
          & Text-to-SQL               &  0.00 /  0.00 / 10.10 & 59.00 / 42.50 / 54.00 &  0.00 /  5.89 / 18.67 &  0.00 /  5.67 / 19.67 \\
          & Text-to-SPARQL            &  0.00 /  0.00 / 14.19 &  0.00 /  7.83 / 17.50 & 99.67 / 60.56 / 69.89 &  0.00 /  4.67 / 14.00 \\
          & Text-to-Cypher            &  0.00 /  0.00 / 18.48 &  0.00 /  8.67 / 18.33 &  0.00 /  6.89 / 29.56 & 79.67 / 56.67 / 66.00 \\
          & KB Routing                & 78.29 / 44.02 / 70.81 & 57.00 / 40.17 / 51.67 & 55.22 / 36.11 / 56.89 & 69.00 / 48.00 / 61.67 \\
          & \cellcolor{citeblue!10}\textbf{OmniRetrieval (Ours)} & \cellcolor{citeblue!10}\textbf{75.33 / 44.09 / 78.14} & \cellcolor{citeblue!10}\textbf{61.00 / 41.50 / 56.17} & \cellcolor{citeblue!10}\textbf{66.67 / 51.22 / 77.22} & \cellcolor{citeblue!10}\textbf{71.33 / 49.67 / 67.33} \\
          \cdashline{2-6}\addlinespace[0.4ex]
          & \textit{Oracle} & \textit{100.00 / 54.51 / 79.19} & \textit{100.00 / 63.83 / 72.50} & \textit{100.00 / 60.89 / 69.78} & \textit{100.00 / 70.67 / 79.33} \\

        \midrule

        \multirow{7}{*}{Gemini-3.1 (Pro)}
          & Document Search           & 90.43 / 52.15 / 78.33 &  0.00 /  2.83 / 13.67 &  0.00 /  0.78 / 52.00 &  0.00 /  4.00 / 19.67 \\
          & Text-to-SQL               &  0.00 /  0.00 /  9.62 & 66.83 / 52.67 / 59.50 &  0.00 /  6.67 / 17.67 &  0.00 /  7.67 / 16.67 \\
          & Text-to-SPARQL            &  0.00 /  0.00 / 15.38 &  0.00 /  7.50 / 12.67 & 100.00 / 71.67 / 79.11 &  0.00 /  3.67 / 14.00 \\
          & Text-to-Cypher            &  0.00 /  0.00 / 10.67 &  0.00 /  8.67 /  9.33 &  0.00 /  7.00 / 16.11 & 81.67 / 53.00 / 61.33 \\
          & KB Routing                & 87.33 / 50.94 / 78.14 & 64.50 / 50.17 / 60.00 & 56.67 / 40.56 / 63.56 & 64.33 / 45.67 / 53.00 \\
          & \cellcolor{citeblue!10}\textbf{OmniRetrieval (Ours)} & \cellcolor{citeblue!10}\textbf{81.19 / 48.65 / 81.52} & \cellcolor{citeblue!10}\textbf{68.67 / 52.33 / 63.67} & \cellcolor{citeblue!10}\textbf{78.67 / 62.11 / 79.89} & \cellcolor{citeblue!10}\textbf{64.67 / 47.67 / 59.33} \\
          \cdashline{2-6}\addlinespace[0.4ex]
          & \textit{Oracle} & \textit{100.00 / 56.13 / 80.24} & \textit{100.00 / 70.00 / 73.67} & \textit{100.00 / 71.44 / 79.78} & \textit{100.00 / 64.67 / 72.33} \\

        \midrule

        \multirow{7}{*}{Sonnet-4.6}
          & Document Search           & 87.62 / 48.33 / 77.43 &  0.00 /  2.33 / 15.17 &  0.00 /  0.11 / 51.33 &  0.00 /  4.00 / 19.33 \\
          & Text-to-SQL               &  0.00 /  0.00 / 17.95 & 64.00 / 47.83 / 58.67 &  0.00 /  6.33 / 22.22 &  0.00 /  7.67 / 19.67 \\
          & Text-to-SPARQL            &  0.00 /  0.00 / 20.52 &  0.00 /  8.50 / 18.17 & 99.22 / 48.89 / 68.56 &  0.00 /  5.00 / 17.00 \\
          & Text-to-Cypher            &  0.00 /  0.00 / 22.90 &  0.00 /  9.17 / 18.17 &  0.00 /  6.89 / 29.00 & 83.33 / 58.67 / 69.33 \\
          & KB Routing                & 77.81 / 42.62 / 72.95 & 60.33 / 44.50 / 56.33 & 40.44 / 19.89 / 58.78 & 63.00 / 47.33 / 59.67 \\
          & \cellcolor{citeblue!10}\textbf{OmniRetrieval (Ours)} & \cellcolor{citeblue!10}\textbf{76.10 / 42.35 / 77.38} & \cellcolor{citeblue!10}\textbf{63.33 / 46.17 / 60.00} & \cellcolor{citeblue!10}\textbf{62.44 / 37.11 / 73.44} & \cellcolor{citeblue!10}\textbf{64.00 / 46.67 / 63.67} \\
          \cdashline{2-6}\addlinespace[0.4ex]
          & \textit{Oracle} & \textit{100.00 / 53.65 / 79.43} & \textit{100.00 / 66.33 / 75.17} & \textit{100.00 / 49.44 / 68.89} & \textit{100.00 / 71.67 / 81.67} \\

        \midrule

        \multirow{7}{*}{Qwen-3.5 (27B)}
          & Document Search           & 85.95 / 44.41 / 71.62 &  0.00 /  3.17 / 12.50 &  0.00 /  1.00 / 45.44 &  0.00 /  4.33 / 15.67 \\
          & Text-to-SQL               &  0.00 /  0.00 /  9.43 & 50.50 / 38.00 / 46.67 &  0.00 /  5.78 / 15.56 &  0.00 /  7.33 / 16.33 \\
          & Text-to-SPARQL            &  0.00 /  0.00 /  8.29 &  0.00 /  8.83 / 10.00 & 99.78 / 52.78 / 62.67 &  0.00 /  5.00 / 15.00 \\
          & Text-to-Cypher            &  0.00 /  0.00 /  8.62 &  0.00 /  8.33 /  8.50 &  0.00 /  6.67 / 13.22 & 81.67 / 59.33 / 67.33 \\
          & KB Routing                & 66.38 / 34.45 / 58.24 & 45.17 / 34.33 / 42.67 & 57.78 / 33.00 / 53.56 & 50.00 / 35.33 / 47.00 \\
          & \cellcolor{citeblue!10}\textbf{OmniRetrieval (Ours)} & \cellcolor{citeblue!10}\textbf{56.14 / 31.12 / 67.86} & \cellcolor{citeblue!10}\textbf{56.00 / 39.67 / 50.33} & \cellcolor{citeblue!10}\textbf{64.78 / 42.22 / 68.44} & \cellcolor{citeblue!10}\textbf{54.33 / 40.33 / 56.67} \\
          \cdashline{2-6}\addlinespace[0.4ex]
          & \textit{Oracle} & \textit{100.00 / 50.72 / 77.00} & \textit{100.00 / 64.17 / 69.50} & \textit{100.00 / 52.89 / 62.56} & \textit{100.00 / 72.67 / 78.67} \\

        \midrule

        \multirow{7}{*}{Gemma-4 (31B)}
          & Document Search           & 85.67 / 44.30 / 72.62 &  0.00 /  3.33 / 17.17 &  0.00 /  0.67 / 47.44 &  0.00 /  4.33 / 20.67 \\
          & Text-to-SQL               &  0.00 /  0.00 / 12.90 & 54.33 / 42.67 / 49.83 &  0.00 /  6.33 / 24.56 &  0.00 /  6.67 / 13.33 \\
          & Text-to-SPARQL            &  0.00 /  0.00 / 10.90 &  0.00 /  9.50 / 10.33 & 98.22 / 57.89 / 65.56 &  0.00 /  4.33 / 16.00 \\
          & Text-to-Cypher            &  0.00 /  0.00 / 13.52 &  0.00 /  9.33 / 10.00 &  0.00 /  6.89 / 16.56 & 76.33 / 52.33 / 60.67 \\
          & KB Routing                & 67.95 / 34.90 / 63.62 & 47.67 / 38.50 / 46.67 & 69.78 / 43.11 / 59.56 & 54.33 / 36.00 / 45.00 \\
          & \cellcolor{citeblue!10}\textbf{OmniRetrieval (Ours)} & \cellcolor{citeblue!10}\textbf{75.00 / 39.38 / 71.48} & \cellcolor{citeblue!10}\textbf{50.50 / 40.83 / 49.83} & \cellcolor{citeblue!10}\textbf{66.44 / 45.00 / 64.56} & \cellcolor{citeblue!10}\textbf{57.67 / 38.67 / 50.67} \\
          \cdashline{2-6}\addlinespace[0.4ex]
          & \textit{Oracle} & \textit{100.00 / 50.19 / 77.48} & \textit{100.00 / 67.33 / 72.00} & \textit{100.00 / 57.89 / 65.78} & \textit{100.00 / 68.00 / 76.00} \\

        \bottomrule
    \end{tabular}
    }
\end{table*}

\clearpage

\begin{figure*}
\centering
\begin{promptbox}
\textbf{[System]}\\[2pt]
You are a query router. Given a question, decide which backend to use (SEARCH, SQL, SPARQL, or CYPHER) and which knowledge base to query. Some queries are ambiguous and may match multiple knowledge bases, return up to <k> routing decisions, most likely first; return fewer if you are confident.

\tcbline

\textbf{[User]}\\[2pt]
\begin{verbatim}
Available knowledge bases:

  SEARCH:
    - <kb_id> [<description> | query type: <...> | examples: <...>]
    - ...
  SQL:
    - <kb_id> [<table names>]
    - ...
  SPARQL:
    - <kb_id> [<description> | examples: <...>]
    - ...
  CYPHER:
    - <kb_id> [nodes: <...> | rels: <...>]
    - ...

Question: <question>

Respond with JSON: {"decisions": [{"route_type": "...", "kb_id": "..."}, ...]}
\end{verbatim}
\end{promptbox}
\caption{Source selection prompt.}
\label{fig:source_selection_prompt}
\end{figure*}

\begin{figure*}
\centering
\begin{promptbox}
\textbf{[System for Text-to-SQL]}\\[2pt]
You are a text-to-SQL translator. Output only the SQL query.

\textbf{[System for Text-to-SPARQL]}\\[2pt]
You are a text-to-SPARQL translator. Output only the SPARQL query.

\textbf{[System for Text-to-Cypher]}\\[2pt]
You are a text-to-Cypher translator. Output only the Cypher query.

\tcbline

\textbf{[User, shared across backends]}\\[2pt]
\begin{verbatim}
<c_b>

Question: <question>

Generate the <SQL|SPARQL|CYPHER> query.
\end{verbatim}
\end{promptbox}
\caption{Query formulation prompts for the structured backends.}
\label{fig:query_formulation_prompt}
\end{figure*}

\begin{figure*}
\centering
\begin{promptbox}
\textbf{[System for Document Search]}\\[2pt]
You are a search query optimizer for a dense retriever. Given the user query and a description of the target corpus, write a hypothetical passage that would be relevant evidence for the query, written in the register, style, and approximate length of documents in that corpus. The passage will be embedded and matched against real corpus documents, so favor concrete, in-domain content over generic phrasing.

Begin your output with the user query verbatim on its own line, just the query text, not the 'Question:' label that precedes it, then on the next line write the hypothetical passage. This keeps the literal query terms in the embedding alongside the semantic expansion.

If the query is a short topic stub or short keyword-style query (just a handful of words, often lowercase and without punctuation), do NOT write a passage, output the verbatim query and nothing else. The bare term already gives a strong dense-retrieval signal, and a hallucinated passage tends to lock onto one specific aspect that may not match the gold document.

Output only the verbatim query followed by the passage (or for a stub query, only the verbatim query), no preamble, no quotes, no labels.
\end{promptbox}
\caption{Query rewriting prompt used in document retrieval. The user prompt template here is the same as for the other backends, shown in Figure~\ref{fig:query_formulation_prompt}.}
\label{fig:search_rewriting_prompt}
\end{figure*}

\begin{figure*}
\centering
\begin{promptbox}
\textbf{[System]}\\[2pt]
You are a result selector. Pick the candidate whose result best answers the question.

\tcbline

\textbf{[User]}\\[2pt]
\begin{verbatim}
Question: <question>

Candidates (each prefixed with its integer index in brackets, e.g. [0], [1], [2]):

[0] <route_type> | <kb_id>
query: <query>
<context and result>

[1] <route_type> | <kb_id>
query: <query>
<context and result>

...

Respond with JSON: {"selected": <integer index>}
\end{verbatim}
\end{promptbox}
\caption{Cross-source evidence selection prompt.}
\label{fig:evidence_selection_prompt}
\end{figure*}

\begin{figure*}
\centering
\begin{promptbox}
\textbf{[System]}\\[2pt]
You are a strict but fair evaluator. You will see a user question and two sides: a PREDICTED side (the model's chosen KB) and a GOLD side (the labeled KB, known to be correct). Each side carries its KB schema/context, the query that was run, and the resulting answer.

Decide whether the predicted side correctly answers the user question. There are two independent ways the prediction can be correct: (1) ANSWER MATCH, the predicted answer is equivalent in meaning to the gold answer, allowing reordering, alias differences, formatting differences, or extra surrounding context; (2) FAITHFUL IMPLEMENTATION ON A DIFFERENT KB, the predicted query faithfully realizes what the user asked, interpreted against the predicted KB schema and data, and the predicted answer is what that query correctly produces. The values may differ entirely from gold because the predicted KB legitimately holds different content. This case applies whenever more than one knowledge base could reasonably answer the same kind of question.

If the gold answer is empty or otherwise degenerate (the gold query may itself be buggy or stale), the gold reference is uninformative, judge by FAITHFUL IMPLEMENTATION alone in that case. Reject when the predicted answer is off-topic or when the predicted query plainly fails to capture what the question is asking. If both pred and gold answers are empty, ALWAYS count it as an ANSWER MATCH (this overrides any reasoning about whether the question should have a real-world answer). If only the predicted answer is empty, reject. Use the schemas and queries on both sides to interpret unfamiliar values.

\tcbline

\textbf{[User]}\\[2pt]
\begin{verbatim}
Question: <question>

[PREDICTED] route=<route> | kb=<kb_id>
query: <query>
<context and answer>

[GOLD] route=<route> | kb=<kb_id>
query: <query>
<context and answer>

Respond with JSON: {"correct": true|false, "reason": "<one-line reason>"}
\end{verbatim}
\end{promptbox}
\caption{LLM-as-a-Judge prompt used to assess answer correctness.}
\label{fig:judge_prompt}
\end{figure*}

}

\end{document}